\documentclass[sigconf,nonacm]{acmart}

\usepackage{algorithm}
\usepackage{algorithmicx}
\usepackage{algpseudocode}
\usepackage{booktabs}
\usepackage{multirow}
\usepackage{amsmath}
\usepackage{graphicx}
\usepackage{hyperref}
\usepackage{xcolor}
\definecolor{vermilion}{RGB}{176,58,46}

\renewcommand\footnotetextcopyrightpermission[1]{} 
\algrenewcommand\algorithmicrequire{\textbf{Input:}}
\algrenewcommand\algorithmicensure{\textbf{Output:}}

\AtBeginDocument{%
  }

\begin{document}

\title{When Surfaces Lie: Exploiting Wrinkle-Induced Attention Shift to Attack Vision-Language Models}

\author{Chengyin Hu, Xuemeng Sun, Jiaju Han, Qike Zhang, Xiang Chen, Xin Wang, Yiwei Wei, Jiahuan Long}


\begin{teaserfigure}
    \centering
    \includegraphics[width=\textwidth]{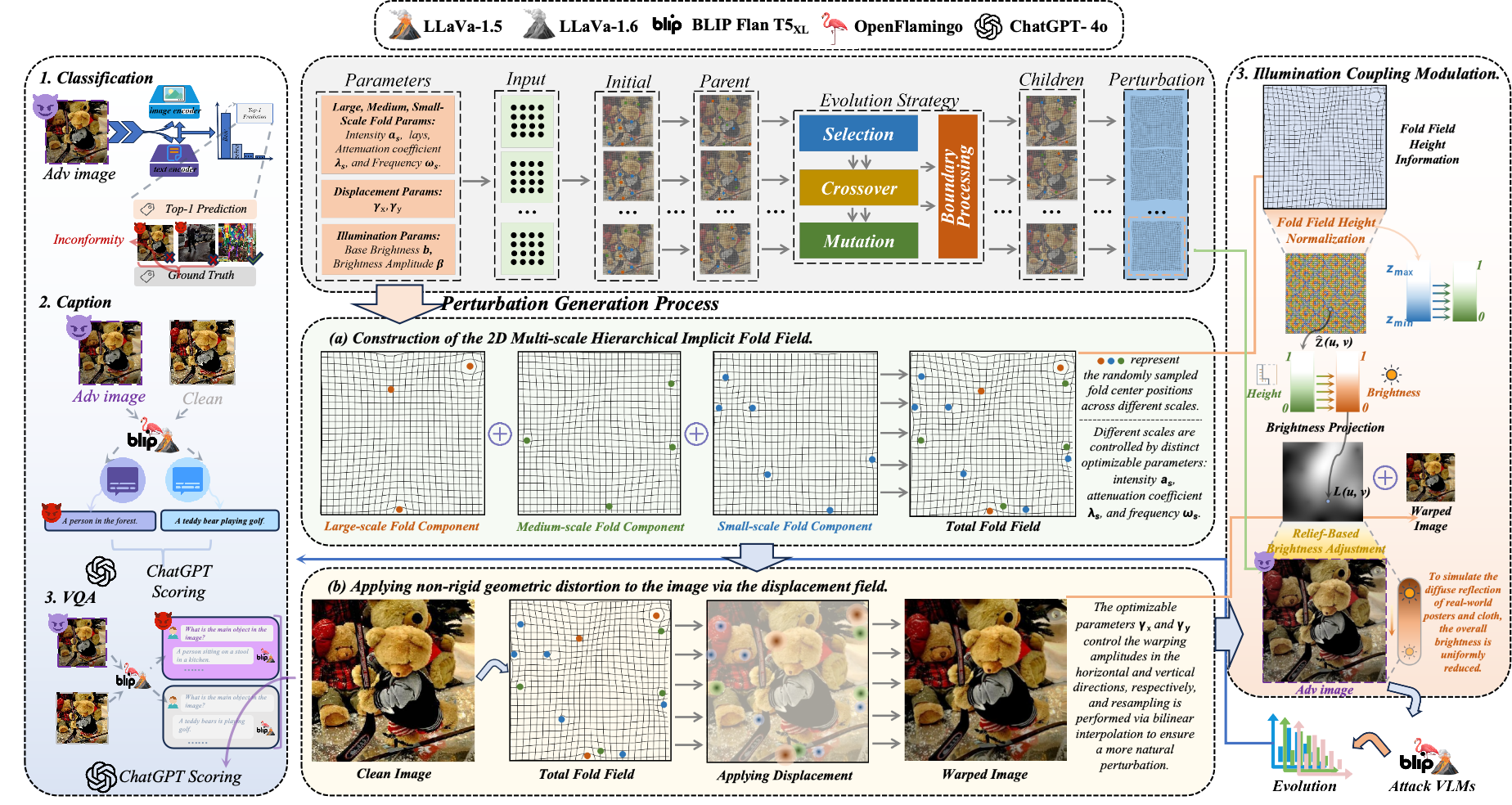}
    \caption{{\bfseries Overview of the Proposed Wrinkle-Based Structural Attack Framework. }{\normalfont The framework models wrinkle-like deformation and uses a genetic algorithm to optimize parameters for generating adversarial examples across vision-language tasks.}}
    \Description{Overview of the proposed wrinkle-based structural attack framework.}
    \label{fig:framework}
\end{teaserfigure}

\begin{abstract}
Visual-Language Models (VLMs) have demonstrated exceptional cross-modal understanding across various tasks, including zero-shot classification, image captioning, and visual question answering. However, their robustness to physically plausible non-rigid deformations-such as wrinkles on flexible surfaces-remains poorly understood. In this work, we propose a parametric structural perturbation method inspired by the mechanics of three-dimensional fabric wrinkles. Specifically, our method generates photorealistic non-rigid perturbations by constructing multi-scale wrinkle fields and integrating displacement field distortion with surface-consistent appearance variations. To achieve an optimal balance between visual naturalness and adversarial effectiveness, we design a hierarchical fitness function in a low-dimensional parameter space and employ an optimization-based search strategy. We evaluate our approach using a two-stage framework: perturbations are first optimized on a zero-shot classification proxy task and subsequently assessed for transferability on generative tasks. Experimental results demonstrate that our method significantly degrades the performance of various state-of-the-art VLMs, consistently outperforming baselines in both image captioning and visual question-answering tasks.
\end{abstract}



\maketitle

\section{Introduction}
Vision-Language Models (VLMs) have emerged as a major paradigm for cross-modal understanding by jointly modeling visual and textual information. Driven by advances in large-scale vision-language pre-training and multimodal instruction tuning, VLMs have demonstrated strong generalization across tasks such as zero-shot classification, image captioning, and visual question answering, and are increasingly regarded as a key route toward open-world visual understanding~\cite{radford2021learning,cherti2023reproducible,li2022blip,dai2023instructblip,alayrac2022flamingo}. Meanwhile, recent progress in test-time generalization, adversarial robustness, and multimodal safety suggests that VLMs are being deployed as increasingly general-purpose systems in complex environments~\cite{cui2024robustness,lu2023set,zhang2022towards,zanella2024test}. Consequently, understanding their robustness under practically grounded distortions has become increasingly important.

However, existing adversarial studies on VLMs still focus mainly on additive pixel-space perturbations, patch attacks, backdoor attacks, and joint image-text manipulations~\cite{carlini2017towards,zhou2022adversarial,zhang2025adversarial}. Comparatively less attention has been paid to non-rigid structural perturbations that arise in realistic imaging conditions. In practice, inputs may be affected by paper creases, surface irregularities, local shadows, lighting changes, and perspective shifts, especially on approximately diffuse surfaces such as paper, posters, cloth, and recaptured screens. Prior work has shown that non-additive factors, including natural shadows, reflected light, illumination variation, physical-world transformations, and differentiable deformation, can also mislead visual models~\cite{athalye2018synthesizing,zhong2022shadows,wang2023rfla,hsiao2024natural,liu2025lighting,lu2024unsegment,ruan2025advdreamer,engstrom2017rotation}. Compared with additive perturbations, structure-aware distortions may reorganize contours, textures, and spatial semantic cues in a qualitatively different way, thereby interfering with cross-modal alignment~\cite{buzhinsky2023metrics}. For vision-language models, such distortions are especially important because they may simultaneously disturb local visual cues and image--text correspondence. Such effects may be amplified in multimodal systems, where errors in visual grounding can further propagate to downstream language outputs. This raises a central question: can wrinkle-like non-rigid structural distortions expose vulnerabilities of VLMs that are not fully revealed by conventional additive or illumination-dominated attacks?

To answer this question, we propose a parametric perturbation framework centered on wrinkle-like distortion. Instead of explicitly simulating complex three-dimensional surface dynamics, our method defines an implicit multi-scale wrinkle field on the image plane to parameterize surface undulations compactly. A displacement field is then derived to impose non-rigid geometric distortion together with mild surface-consistent appearance variation. In this way, the perturbation is modeled primarily as a structured deformation process rather than a direct pixel-space modification, providing a compact approximation to wrinkle-like imaging distortions in practical scenarios. Because the perturbation is controlled by low-dimensional variables, including wrinkle intensity, layer number, attenuation coefficient, frequency, displacement amplitude, and related appearance parameters, the attack can be formulated as a black-box optimization problem in parameter space. To balance attack effectiveness and visual plausibility, we further design a hierarchical fitness function and employ a genetic algorithm~\cite{holland1992adaptation,alzantot2019genattack} to search for effective perturbation parameters.

Our goal is to examine whether wrinkle-like structural distortion defines a distinct and practically meaningful robustness setting for multimodal systems, rather than simply introducing another perturbation type for VLMs. Unlike pixel-level perturbations, it locally reorganizes spatial layout and may therefore simultaneously affect object boundaries, local semantic cues, and image--text alignment. We evaluate the proposed perturbation on zero-shot classification across multiple CLIP models, and then transfer the generated adversarial examples to image captioning and visual question answering. Through this unified pipeline, we systematically investigate the vulnerability of VLMs under wrinkle-like structural perturbations with mild surface-consistent appearance variation. The main contributions of this paper are as follows:
\begin{itemize}
    \item We propose a wrinkle-like structural attack for VLMs based on multi-scale wrinkle modeling, displacement-field warping, and surface-consistent appearance modulation.

   \item We conduct extensive experiments on zero-shot classification, image captioning, and visual question answering, showing consistent performance degradation across multiple state-of-the-art vision-language models and stronger attack effectiveness than representative baselines.

    \item We perform ablation studies on search hyperparameters, multi-scale design, and perceptual weighting, revealing multi-scale complementarity and the trade-off between adversarial effectiveness and visual naturalness.
\end{itemize}

\section{Related Work}
\subsection{Adversarial Vulnerabilities in VLMs}
With the rapid development of vision-language pre-trained Models and large-scale multimodal models, their security in open-world scenarios has attracted increasing attention~\cite{zhou2022adversarial,zhang2025adversarial}. Existing studies show that visual-side adversarial perturbations can substantially degrade VLM performance on image classification, image captioning, and visual question answering, and may further affect textual outputs through cross-modal alignment~\cite{cui2024robustness}. Subsequent work has investigated transferable attacks against vision-language pre-training models, showing that cross-modal systems can still be misled without access to target-model gradients~\cite{lu2023set}. More recent studies further extend this line to stronger and sequential transfer attacks, suggesting that VLM vulnerabilities are not limited to a single task but can generalize across multiple downstream settings~\cite{xie2025chain}.

\subsection{Structured and Physical Perturbations}
Compared with pixel-level additive perturbations, structured and physically plausible adversarial disturbances better resemble practical imaging conditions and therefore have greater real-world relevance~\cite{carlini2017towards}. Previous studies have shown that natural shadow, reflected light, illumination variation, and deformation can all mislead visual models~\cite{zhong2022shadows,wang2023rfla,hsiao2024natural,liu2025lighting,lu2024unsegment,ruan2025advdreamer}. For approximately diffuse surfaces, classical reflectance analysis further suggests that changes in surface orientation may induce systematic luminance variation under illumination, providing a physical basis for mild appearance changes associated with wrinkle deformation~\cite{basri2003lambertian}. These findings indicate that effective real-world attacks need not appear as obvious noise, but may instead manifest as structural distortion, illumination change, or surface undulation. However, existing studies mainly emphasize shadow, lighting, or deformation-based perturbations, while wrinkle-like non-rigid distortions remain largely unexplored in current VLM robustness research~\cite{zhong2022shadows,hsiao2024natural,liu2025lighting,lu2024unsegment,ruan2025advdreamer}.

\section{Methodology}
Figure~\ref{fig:framework} illustrates the overall framework of the proposed method. We first model wrinkle-like structural perturbations with a multi-scale fold field, convert them into non-rigid geometric warping, and apply a mild appearance adjustment to obtain the final adversarial examples. The perturbation parameters are optimized in a low-dimensional black-box space using a genetic algorithm. The resulting adversarial examples are evaluated on zero-shot classification and further transferred to image captioning and VQA tasks.

\subsection{Problem Definition}
This section introduces the proposed wrinkle-like structural perturbation method. Our objective is to assess the robustness of vision-language models under practically motivated structural distortions. Given an input image $x \in \mathbb{R}^{H \times W \times C}$ and its ground-truth label $y$, we aim to generate a perturbed image $x'$ that changes the model prediction while preserving visual plausibility as much as possible.

In our experiments, the perturbation generation process is modeled as a stochastic parameterized mapping:
\begin{equation}
x' = T_{\theta}(x;\xi),
\end{equation}
where $\theta$ denotes the perturbation parameters to be optimized, mainly including the intensity, number of layers, decay rate, frequency, displacement magnitude, and related appearance parameters of wrinkles at different scales; $\xi$ denotes the random variables involved in perturbation generation, corresponding to the random sampling of wrinkle center locations at each scale. Since paper bending, surface undulation, and local distortion in practical scenes usually exhibit a certain degree of randomness, we model the perturbation as a stochastic parameterized non-rigid structural perturbation.

For each zero-shot classification model, we adopt an un-targeted attack setting. Let $f : \mathbb{R}^{H \times W \times C} \rightarrow \mathbb{R}^{K}$ denote the classifier, and let:
\begin{equation}
\hat{y}(x')=\arg\max_{c\in\{1,\dots,K\}} f_{c}(x').
\end{equation}
where $f_{c}(x')$ denotes the logit (or confidence score) of class $c$. The attack objective is to find a perturbed image $x'=T_{\theta}(x;\xi)$ such that:
\begin{equation}
\hat{y}(x') \neq y.
\end{equation}
The perturbation parameters are optimized independently for each zero-shot classification model. The adversarial examples generated in this stage are then transferred to image captioning and visual question answering to evaluate the cross-task transferability of this type of structural perturbation on generative vision-language tasks.

\subsection{Multi-Scale Wrinkle-Like Distortion}
To efficiently simulate wrinkle-like non-rigid deformation, we model surface undulation on the image plane with an implicit multi-scale height field.

Let the normalized image coordinates be $(u, v)$. The total wrinkle field is defined as:
\begin{equation}
z(u,v)=z_L(u,v)+z_M(u,v)+z_S(u,v).
\end{equation}
Here, $z_L(u,v)$, $z_M(u,v)$, and $z_S(u,v)$ denote the large-scale, medium-scale, and small-scale wrinkle components, respectively. The large-scale component captures the overall bending trend, the medium-scale component describes local hierarchical variation, and the small-scale component supplements fine-grained surface details.

For any scale $s \in \{L, M, S\}$, the corresponding wrinkle component is constructed by superposing multiple local wrinkle sources:
\begin{equation}
z_s(u,v)=\sum_{k=1}^{K_s} a_s \cdot \phi_s(d_{s,k}(u,v)) \cdot \exp\!\left(-\lambda_s \cdot d_{s,k}(u,v)\right),
\end{equation}
where $K_s$ is the number of wrinkle layers at scale $s$, $a_s$ is the wrinkle intensity, $\lambda_s$ is the decay coefficient, and $d_{s,k}(u,v)$ is the distance from position $(u,v)$ to the $k$-th wrinkle center:
\begin{equation}
d_{s,k}(u,v)=\sqrt{(u-c^{u}_{s,k})^2+(v-c^{v}_{s,k})^2},
\end{equation}
where $(c^{u}_{s,k},c^{v}_{s,k})$ denotes the randomly sampled center of the corresponding wrinkle source. Since the spatial distributions of wrinkle centers differ across scales, the resulting field can represent both the overall deformation trend and local structural complexity.

In implementation, different oscillation functions are used at different scales. The large-scale and small-scale components are modeled with sine functions:
\begin{equation}
\phi_L(d)=\sin(\omega_L\pi d),\quad \phi_S(d)=\sin(\omega_S\pi d),
\end{equation}
while the medium-scale component uses a combination of sine and cosine functions to enrich mid-level structural variation:
\begin{equation}
\phi_M(d)=\tfrac{1}{2}\left[\sin(\omega_{M1}\pi d)+\cos(\omega_{M2}\pi d)\right].
\end{equation}
This formulation gives the wrinkle field clear multi-scale hierarchical characteristics. Compared with directly adding independent perturbations in pixel space, it provides a more structured approximation to surface bending, cloth undulation, and local non-rigid deformation in practical scenes.

\textbf{Displacement field construction and image warping.} After generating the wrinkle field $z(u,v)$, we construct a two-dimensional displacement field from its local variations. Intuitively, regions with stronger wrinkle variation should induce stronger stretching, compression, or bending in the corresponding image content. Following common practices in spatial transformation and deformation modeling~\cite{jaderberg2015spatial,ruan2025advdreamer}, the displacement field is approximated as:
\begin{equation}
r_u(u,v)=\gamma_u \cdot \frac{\partial z(u,v)}{\partial u},\quad
r_v(u,v)=\gamma_v \cdot \frac{\partial z(u,v)}{\partial v},
\end{equation}
where $\gamma_u$ and $\gamma_v$ control the distortion magnitudes in the horizontal and vertical directions, respectively. The position with coordinates $(u,v)$ in the original image is then mapped to a new sampling position:
\begin{equation}
\hat{u}=u+r_u(u,v),\quad \hat{v}=v+r_v(u,v).
\end{equation}
The original image is then resampled by bilinear interpolation to obtain the geometrically distorted image:
\begin{equation}
x_w = W(x,r_u,r_v),
\end{equation}
where $W(\cdot)$ denotes the image resampling operator based on the displacement field. Rather than modifying colors pixel by pixel, this process reorganizes the spatial layout of image content according to the wrinkle field, producing non-rigid deformation in contours, local structures, and fine-scale details. Compared with additive perturbations, this distortion pattern is more consistent with imaging variations caused by paper bending, surface undulation, or cloth wrinkles in practical scenarios.

\textbf{Surface-consistent appearance adjustment.} In practical scenes, surface undulations may induce mild brightness variation in addition to local geometric changes. To account for this effect, we introduce a lightweight appearance adjustment module after geometric distortion, inspired by illumination-related adversarial perturbations and physical rendering variation~\cite{hsiao2024natural,liu2025lighting}. Specifically, image brightness is adjusted according to the wrinkle field distribution. The wrinkle field is normalized as:
\begin{equation}
\hat{z}(u,v)=\frac{z(u,v)-z_{\min}}{z_{\max}-z_{\min}+\epsilon_z}
\end{equation}
where $z_{\min}$ and $z_{\max}$ denote the minimum and maximum values of the current wrinkle field, respectively, and $\epsilon_z$ is a small constant to prevent division by zero. The brightness modulation map is then constructed as:
\begin{equation}
L(u,v)=b+\beta\cdot \hat{z}(u,v),
\end{equation}
where $b$ denotes the base brightness and $\beta$ denotes the amplitude of brightness variation. The final perturbed image is expressed as:
\begin{equation}
x'=\Pi_{[0,1]}(x_w \odot L),
\end{equation}
where $\odot$ denotes element-wise multiplication, and $\Pi_{[0,1]}(\cdot)$ denotes projecting the input element-wise onto the interval $[0,1]$ so that the output pixel values remain within the valid range. Under this formulation, the final perturbation is modeled primarily as wrinkle-like structural deformation with mild surface-consistent appearance variation, rather than as a standalone illumination attack.

\textbf{Perceptual similarity constraint.} To balance attack effectiveness and visual plausibility, we define a perceptual similarity constraint to measure the visual closeness between the perturbed image and the original image. Since a single metric cannot fully capture both structural preservation and high-level perceptual consistency, we combine SSIM~\cite{wang2004image} and LPIPS~\cite{zhang2018unreasonable} as
\begin{equation}
S_{\mathrm{perc}}(x,x')=(1-\alpha_1)\cdot S_{\mathrm{SSIM}}(x,x')+\alpha_1\cdot \bigl(1-S_{\mathrm{LPIPS}}(x,x')\bigr),
\end{equation}
where $\alpha_1 \in [0,1]$ controls the trade-off between perceptual consistency and structural preservation. Here, $S_{\mathrm{SSIM}}$ measures local structural fidelity, while $1-S_{\mathrm{LPIPS}}(x,x')$ serves as a similarity-oriented perceptual term.

\textbf{Fitness function.} For each zero-shot classification model, we define the attack objective as reducing the model's prediction confidence for the true class. Let the predicted probability of the perturbed image $x'$ on the true class $y$ be $p(y\mid x')$. Then, the adversarial score is defined as:
\begin{equation}
S_{\mathrm{adv}}(x')=-\log(p(y\mid x')+\epsilon_{\mathrm{adv}}),
\end{equation}
where $\epsilon_{\mathrm{adv}}$ is a numerical stability term. For unified modeling with the perceptual similarity term, the attack score is further normalized and denoted as $S_{\mathrm{ladv}}\in[0,1]$.

On this basis, we adopt a hierarchical fitness design that explicitly prioritizes attack success. First, the attack success indicator is defined as:
\begin{equation}
I(x')=
\begin{cases}
1, & \hat{y}(x') \neq y,\\
0, & \hat{y}(x') = y.
\end{cases}
\end{equation}
$I(x')$ selects the branch of the hierarchical fitness. When the perturbation has not yet caused misclassification, both attack effectiveness and perceptual quality are considered, and the fitness is defined as:
\begin{equation}
F(\theta,\xi)=(1-\alpha_2) S_{\mathrm{ladv}}+\alpha_2 S_{\mathrm{perc}},
\end{equation}
where $\alpha_2 \in [0,1]$ balances attack effectiveness and perceptual quality. When the perturbation has already changed the model prediction, the fitness further favors candidates with higher perceptual quality among successful samples, and is defined as:
\begin{equation}
F(\theta,\xi)=1+\eta S_{\mathrm{perc}},
\end{equation}
where $\eta \in [0,1]$ controls the preference for perceptual quality among successful samples.

During the search, attack success is treated as the primary objective, after which the perceptual quality of the perturbed image is further optimized. Since both $S_{\mathrm{ladv}}$ and $S_{\mathrm{perc}}$ are normalized to $[0,1]$, the fitness values of unsuccessful samples lie in $[0,1]$, whereas those of successful samples are greater than 1, thereby explicitly separating successful and failed candidates. This design avoids overly large empirical constants and is more consistent with fitness modeling in genetic algorithms under black-box optimization.

\subsection{Genetic Algorithm Optimization}

\begin{algorithm}[H]
\caption{Pseudocode of Our Method}
\label{alg:attack_l1}
\begin{algorithmic}[1]
\Require original image $x$, true label $y$, zero-shot classification model $f$, parameter space $\Theta$, population size $N$, maximum generations $G$, elite size $E$, mutation rate $p_m$, stagnation threshold $K$
\Ensure optimal perturbation parameters $\theta^\ast$, perturbed image $x^{\prime\ast}$
\For{$t=0$ to $G-1$}
    \For{each $\theta_i^{(t)} \in P^{(t)}$}
        \State $x'_i \gets T_{\theta_i^{(t)}}(x;\xi_i)$
        \State $S_{\mathrm{adv},i} \gets S_{\mathrm{adv}}(x'_i)$
        \State $S_{\mathrm{perc},i} \gets S_{\mathrm{perc}}(x,x'_i)$
        \State $F_i^{(t)} \gets F(\theta_i^{(t)},\xi_i)$
    \EndFor
    \State $P_{\mathrm{elite}}^{(t)} \gets \mathrm{TopE}(P^{(t)},F^{(t)})$
    \State $Q^{(t)} \gets \mathrm{Select}(P^{(t)},F^{(t)})$
    \State $\tilde{P}^{(t)} \gets C(Q^{(t)})$
    \State $\hat{P}^{(t)} \gets M(\tilde{P}^{(t)},p_m)$
    \State $P_{\mathrm{rand}}^{(t)} \sim U(\Theta)$
    \State $P^{(t+1)} \gets P_{\mathrm{elite}}^{(t)} \cup \hat{P}^{(t)} \cup P_{\mathrm{rand}}^{(t)}$
    \If{stagnation for $K$ generations}
        \State partially reinitialize $P^{(t+1)} \setminus P_{\mathrm{elite}}^{(t)}$
    \EndIf
\EndFor
\State $\theta^\ast \gets \arg\max_{\theta_i^{(t)}} F_i^{(t)}$
\State $x^{\prime\ast} \gets T_{\theta^\ast}(x;\xi^\ast)$
\State \Return $\theta^\ast, x^{\prime\ast}$
\end{algorithmic}
\end{algorithm}
Since perturbation generation involves random wrinkle center sampling, geometric resampling, and mild appearance adjustment, the resulting objective is non-convex, nonlinear, and stochastic, making stable optimization difficult for traditional gradient-based methods. Moreover, we consider a black-box attack setting and do not rely on gradient information from the target model. Therefore, a Genetic Algorithm (GA)~\cite{holland1992adaptation} is adopted to search in the low-dimensional parameter space. The pseudocode of the proposed method is given in Algorithm \ref{alg:attack_l1}. The method takes the original image, the ground-truth label, the zero-shot classification surrogate model, and the genetic search parameters as inputs, and outputs the optimal perturbation parameters together with the corresponding adversarial example.

\section{Experiments}
\subsection{Experimental Setup}
To evaluate the proposed wrinkle-like structural perturbation, we conduct experiments on zero-shot classification, image captioning, and visual question answering. In zero-shot classification, four CLIP-based models are independently used to generate and optimize perturbations under the same black-box setting. The adversarial examples are then transferred to image captioning and visual question answering to evaluate cross-task transferability.

\textbf{Dataset.}
To ensure consistent comparison across the three tasks, all experiments are conducted on a unified data source, reducing the influence of cross-dataset distribution differences. Specifically, we select 30 categories with 10 images per category from the COCO dataset~\cite{lin2014microsoft}, resulting in 300 clean instance images. For comparison, we follow the same COCO-based evaluation setting as the reference study to keep the data source aligned with the reported baselines~\cite{liu2025lighting}.

\textbf{Model Settings.}
For zero-shot classification, we adopt four CLIP-based models with different visual encoders, including OpenCLIP ViT-B/16~\cite{cherti2023reproducible}, Meta-CLIP ViT-L/14~\cite{xu2023demystifying,chuang2025meta}, EVA-CLIP ViT-G/14~\cite{sun2023eva}, and OpenAI CLIP ViT-L/14~\cite{radford2021learning}. For image captioning and visual question answering, we use generative vision-language models for transfer evaluation, including LLaVA-1.5~\cite{liu2023visual}, LLaVA-1.6~\cite{liu2024improved}, OpenFlamingo~\cite{awadalla2023openflamingo}, Blip-2 ViT-L~\cite{li2022blip}, Blip-2 FlanT5-XL~\cite{li2023blip}, and InstructBLIP~\cite{dai2023instructblip}. The baseline methods include Shadow Attack~\cite{zhong2022shadows}, Natural Light Attack~\cite{hsiao2024natural}, and ITA~\cite{liu2025lighting}. All methods are evaluated under the same protocol for fair comparison.

\textbf{Evaluation Metrics.}
For zero-shot classification, classification accuracy (ACC) is used as the primary metric in the main comparison, while attack success rate (ASR) is additionally reported in the ablation study to reflect search effectiveness. Here, ACC denotes the classification accuracy on the perturbed test set, and ASR denotes the proportion of samples whose predicted labels are altered by the perturbation. We further report a GPT-4o-based visual naturalness score to evaluate the perceptual realism of perturbed images in zero-shot classification. Higher GPT-4o scores indicate better visual naturalness, and all methods are evaluated under the same scoring prompt and protocol. For image captioning, we adopt a judge-based evaluation protocol and use GPT-4 to assess the semantic consistency between generated captions and image content. For visual question answering, the generated answer and the ground-truth answer are jointly provided to GPT-4, which determines whether the predicted answer is correct. This protocol follows recent robustness studies on large multimodal models and is adopted here to ensure fair comparison under a shared setting~\cite{cui2024robustness,liu2025lighting,verma2024evaluating}. For fair comparison with prior baselines, we follow the reference study in both the COCO-based data setting and the GPT-based evaluation protocol.

\textbf{Experimental Parameters.}
In our experiments, wrinkle-like perturbations are searched in a low-dimensional parameter space using a genetic algorithm under a black-box setting. The population size is set to 8, the maximum number of fitness evaluations for each image is set to 350 as the evaluation budget, the perceptual fusion coefficient $\alpha_1$ in the perceptual similarity term is set to 0.7, and the perceptual weighting coefficient $\alpha_2$ in Eq.~(18) is set to 0.3. The perturbation parameters include wrinkle-field generation parameters, displacement parameters, and related appearance parameters. Specifically, the strength ranges of the large- and medium-scale wrinkle components are set to $[0.4, 0.8]$, while that of the small-scale component is set to $[0.3, 0.5]$; the layer numbers of the large-, medium-, and small-scale components are set to $[2, 4]$, $[4, 6]$, and $[6, 8]$, respectively. The decay and frequency parameters are searched within preset ranges for different scales. The displacement parameters $\gamma_u$ and $\gamma_v$ are both set to $[0.4, 0.6]$, while the base brightness and brightness amplitude are set to $[0.4, 0.8]$ and $[0.2, 0.4]$, respectively. In the generative tasks, image captioning and visual question answering directly use the adversarial examples generated in the zero-shot classification stage for transfer evaluation. All experiments are conducted on an RTX 5090 GPU.

\begin{table*}[t]
  \centering
  \setlength{\tabcolsep}{4.8pt}
  \renewcommand{\arraystretch}{1.12}
  \caption{{\bfseries Performance of VLMs on Zero-shot Classification under Wrinkle-like Structural Perturbations Generated from the COCO Dataset.} {\normalfont We report both classification accuracy (ACC) and GPT-4o-based visual naturalness scores. Lower ACC indicates stronger attack effectiveness, while higher GPT-4o scores indicate better visual naturalness. The best results are shown in bold, and the second-best results are underlined.}}
  \label{tab:acc_comparison}
  \resizebox{\textwidth}{!}{
  \begin{tabular}{lcccccccc}
    \toprule
    \multirow{2}{*}{Method}
      & \multicolumn{2}{c}{OpenCLIP ViT-B/16}
      & \multicolumn{2}{c}{Meta-CLIP ViT-L/14}
      & \multicolumn{2}{c}{EVA-CLIP ViT-G/14}
      & \multicolumn{2}{c}{OpenAI CLIP ViT-L/14} \\
    \cmidrule(lr){2-3} \cmidrule(lr){4-5} \cmidrule(lr){6-7} \cmidrule(lr){8-9}
    & ACC (\%) & GPT-4o
    & ACC (\%) & GPT-4o
    & ACC (\%) & GPT-4o
    & ACC (\%) & GPT-4o \\
    \midrule
    Clean
      & 97 & 3.65
      & 98 & 3.65
      & 98 & 3.65
      & 93 & 3.65 \\

    Natural Light Attack~\cite{hsiao2024natural}
      & 94 ($\downarrow 3$) & \underline{2.32} ($\downarrow 1.33$)
      & 97 ($\downarrow 1$) & 2.14 ($\downarrow 1.51$)
      & 97 ($\downarrow 1$) & 2.16 ($\downarrow 1.49$)
      & 93 (--)
      & 2.07 ($\downarrow 1.58$) \\

    Shadow Attack~\cite{zhong2022shadows}
      & 84 ($\downarrow 13$) & 2.11 ($\downarrow 1.54$)
      & 82 ($\downarrow 16$) & 1.91 ($\downarrow 1.74$)
      & 95 ($\downarrow 3$) & 2.03 ($\downarrow 1.62$)
      & 79 ($\downarrow 14$) & 1.78 ($\downarrow 1.87$) \\

    ITA~\cite{liu2025lighting}
      & \underline{46} ($\downarrow 51$) & 2.11 ($\downarrow 1.54$)
      & \underline{64} ($\downarrow 34$) & 2.15 ($\downarrow 1.50$)
      & 84 ($\downarrow 14$) & 2.17 ($\downarrow 1.48$)
      & \underline{51} ($\downarrow 42$) & 2.15 ($\downarrow 1.50$) \\

    \addlinespace[2pt]

    Ours (random)
      & 52 (\textcolor{vermilion}{$\downarrow 45$}) & \textbf{2.71} (\textcolor{vermilion}{$\downarrow 0.94$})
      & 68 (\textcolor{vermilion}{$\downarrow 30$}) & \textbf{2.47} (\textcolor{vermilion}{$\downarrow 1.18$})
      & \underline{75} (\textcolor{vermilion}{$\downarrow 23$}) & \textbf{2.89} (\textcolor{vermilion}{$\downarrow 0.76$})
      & 53 (\textcolor{vermilion}{$\downarrow 40$}) & \textbf{2.64} (\textcolor{vermilion}{$\downarrow 1.01$}) \\

    Ours (best)
      & \textbf{43} (\textcolor{vermilion}{$\downarrow 54$}) & 2.29 (\textcolor{vermilion}{$\downarrow 1.36$})
      & \textbf{58} (\textcolor{vermilion}{$\downarrow 40$}) & \underline{2.39} (\textcolor{vermilion}{$\downarrow 1.26$})
      & \textbf{66} (\textcolor{vermilion}{$\downarrow 32$}) & \underline{2.72} (\textcolor{vermilion}{$\downarrow 0.93$})
      & \textbf{42} (\textcolor{vermilion}{$\downarrow 51$}) & \underline{2.51} (\textcolor{vermilion}{$\downarrow 1.14$}) \\
    \bottomrule
  \end{tabular}}
\end{table*}

\begin{figure*}[t]
    \centering
    \begin{minipage}[t]{0.48\textwidth}
        \centering
        \includegraphics[width=\linewidth]{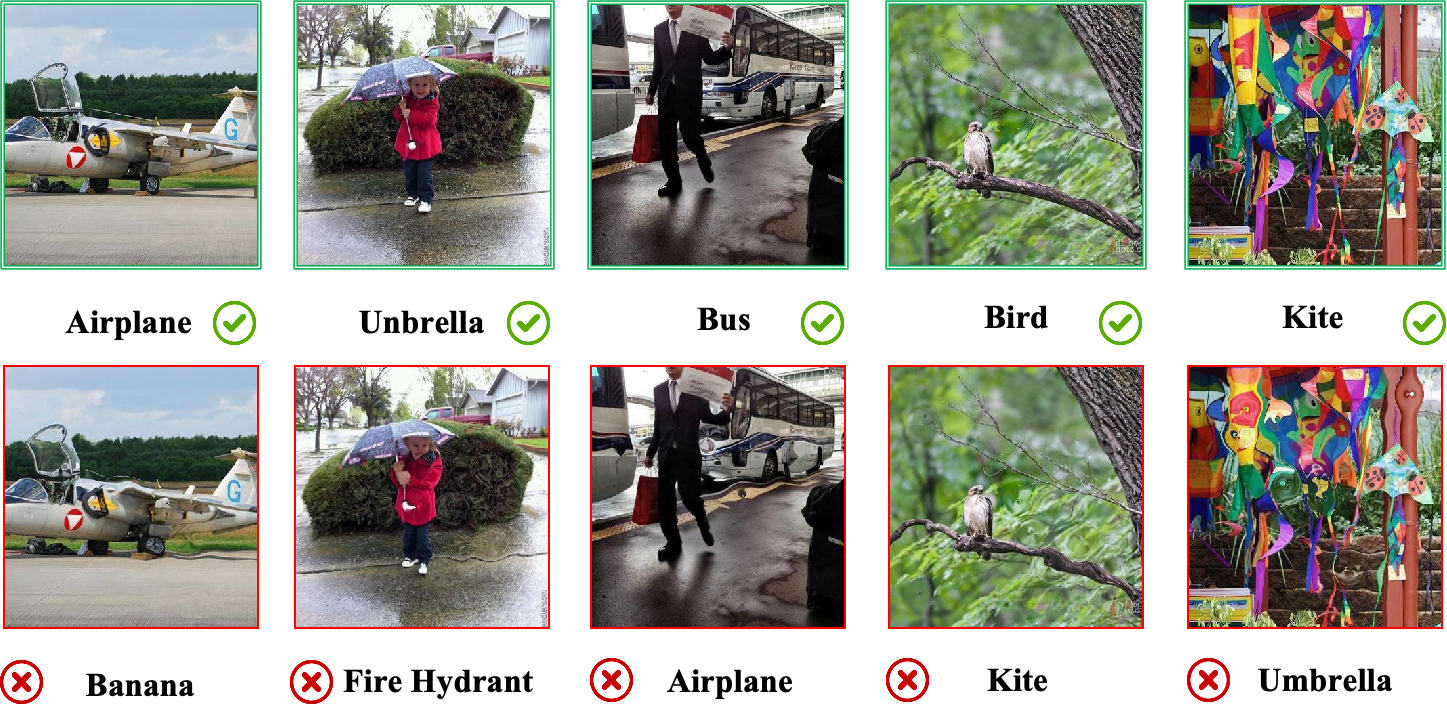}
        \caption*{(a)}
    \end{minipage}\hfill
    \begin{minipage}[t]{0.48\textwidth}
        \centering
        \includegraphics[width=\linewidth]{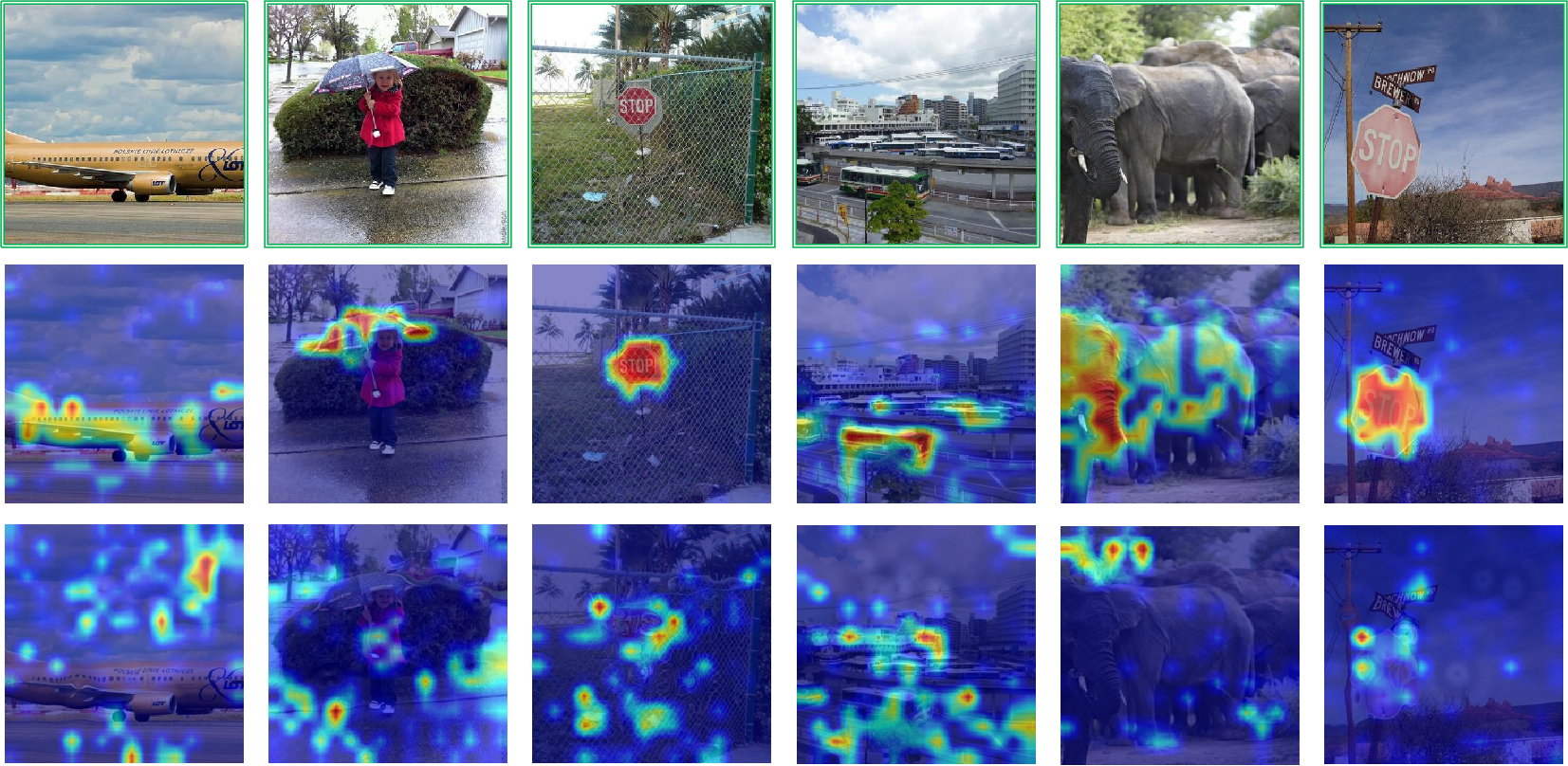}
        \caption*{(b)}
    \end{minipage}
    \vspace{-5pt}\caption{{\bfseries Qualitative Comparison Between Clean and Adversarial Examples. }{\normalfont (a) Representative clean and adversarial image pairs, where the proposed perturbation remains visually subtle but causes erroneous predictions on adversarial samples. (b) Comparison of attention visualizations between clean and adversarial images, showing that the perturbation redirects model attention away from semantically meaningful regions.}}
    \label{fig:figure2}
\end{figure*}

\subsection{Comparative Experiments}
\noindent\textbf{Zero-shot Classification Experiments.}
We first evaluate the proposed wrinkle-like structural perturbation on zero-shot classification to examine whether it can effectively degrade the recognition performance of CLIP models. For each CLIP model, perturbation parameters are independently optimized using the genetic algorithm, and the resulting adversarial examples are evaluated on the same source model.

As shown in Table~\ref{tab:acc_comparison}, the proposed method achieves the strongest attack performance across all four CLIP models in terms of classification accuracy. Compared with clean samples, \emph{Ours (best)} reduces the accuracy of OpenCLIP ViT-B/16, Meta-CLIP ViT-L/14, EVA-CLIP ViT-G/14, and OpenAI CLIP ViT-L/14 by 54, 40, 32, and 51 percentage points, respectively, consistently outperforming Natural Light Attack~\cite{hsiao2024natural}, Shadow Attack~\cite{zhong2022shadows}, and ITA~\cite{liu2025lighting}. Meanwhile, \emph{Ours (random)} achieves the highest GPT-4o visual naturalness scores on all four models, indicating that the proposed perturbation can better preserve perceptual realism while maintaining substantial attack effectiveness. Notably, this advantage is not limited to a single backbone, but remains consistent across CLIP variants with different architectures and scales. This suggests that the proposed perturbation captures a transferable structural vulnerability in visual encoding rather than exploiting a model-specific weakness. These results indicate that the proposed perturbation achieves a favorable trade-off between attack strength and visual naturalness.

\begin{table*}[t]
  \centering
  \caption{{\bfseries Performance Degradation ($\downarrow$) of VLMs on Image Captioning under Wrinkle-like Structural Perturbations Generated from the COCO Dataset.} {\normalfont We compare the semantic consistency scores (\%) evaluated by GPT-4 for baseline methods and the proposed method. Lower values indicate stronger attack performance. Numbers in bold denote the best results, and underlined numbers denote the second-best results.}}
  \label{tab:transfer_zeroshot}
  \footnotesize
  \setlength{\tabcolsep}{3.2pt}
  \renewcommand{\arraystretch}{1.0}
  \begin{tabular}{llccccccc}
    \toprule
    Image Encoder & Models & Params & Clean & Natural Light Attack~\cite{hsiao2024natural} & Shadow Attack~\cite{zhong2022shadows} & ITA~\cite{liu2025lighting} & Ours (random) & Ours (best) \\
    \midrule
    \multirow{4}{*}{OpenAI CLIP ViT-L/14}
      & LLaVA-1.5 & 7B & 78.60 & 77.00 ($\downarrow 1.63$) & 74.60 ($\downarrow 4.00$) & 63.73 ($\downarrow 14.87$) & \underline{56.58} (\textcolor{vermilion}{$\downarrow 22.02$}) & \textbf{54.51} (\textcolor{vermilion}{$\downarrow 24.09$}) \\
      & LLaVA-1.6 & 7B & 72.10 & 71.70 ($\downarrow 0.39$) & 71.17 ($\downarrow 0.93$) & 61.60 ($\downarrow 10.51$) & \underline{54.45} (\textcolor{vermilion}{$\downarrow 17.65$}) & \textbf{53.15} (\textcolor{vermilion}{$\downarrow 18.95$}) \\
      & OpenFlamingo & 3B & 70.20 & 69.53 ($\downarrow 0.67$) & 67.80 ($\downarrow 2.40$) & 53.93 ($\downarrow 16.27$) & \underline{33.59} (\textcolor{vermilion}{$\downarrow 36.61$}) & \textbf{30.18} (\textcolor{vermilion}{$\downarrow 50.02$}) \\
      & Blip-2 (FlanT5\textsubscript{XL}, ViT-L) & 3.4B & 75.10 & 70.77 ($\downarrow 4.33$) & 68.57 ($\downarrow 6.53$) & \underline{60.93} ($\downarrow 14.27$) & 61.01 (\textcolor{vermilion}{$\downarrow 14.09$}) & \textbf{58.21} (\textcolor{vermilion}{$\downarrow 16.89$}) \\
    \midrule
    \multirow{2}{*}{EVA-CLIP ViT-G/14}
      & Blip-2 (FlanT5\textsubscript{XL}) & 4.1B & 74.96 & 71.27 ($\downarrow 3.69$) & 68.80 ($\downarrow 6.17$) & 62.01 ($\downarrow 11.88$) & \underline{59.48} (\textcolor{vermilion}{$\downarrow 15.48$}) & \textbf{57.08} (\textcolor{vermilion}{$\downarrow 17.88$}) \\
      & InstructBLIP (FlanT5\textsubscript{XL}) & 4.1B & 76.50 & 72.07 ($\downarrow 4.43$) & 69.77 ($\downarrow 6.73$) & \underline{63.20} ($\downarrow 13.30$) & 63.92 (\textcolor{vermilion}{$\downarrow 12.58$}) & \textbf{56.57} (\textcolor{vermilion}{$\downarrow 19.93$}) \\
    \bottomrule
  \end{tabular}
\end{table*}

\begin{table*}[t]
  \centering
  \caption{{\bfseries Performance Degradation ($\downarrow$) of VLMs on the VQA task under Wrinkle-like Structural Perturbations Generated from the COCO Dataset.} {\normalfont We compare the answer correctness scores (\%) evaluated by GPT-4 for baseline methods and the proposed method. Lower values indicate stronger attack performance. Bold numbers denote the best results, and underlined numbers denote the second-best results.}}
  \label{tab:vqa}
  \footnotesize
  \setlength{\tabcolsep}{3.2pt}
  \renewcommand{\arraystretch}{1.0}
  \begin{tabular}{llccccccc}
    \toprule
    Image Encoder & Models & Params & Clean & Natural Light Attack~\cite{hsiao2024natural} & Shadow Attack~\cite{zhong2022shadows} & ITA~\cite{liu2025lighting} & Ours (random) & Ours (best) \\
    \midrule
    \multirow{4}{*}{OpenAI CLIP ViT-L/14}
      & LLaVA-1.5 & 7B & 68.00 & 68.00 ($\downarrow 0.00$) & 67.00 ($\downarrow 1.00$) & \underline{48.00} ($\downarrow 20.00$) & 55.00 (\textcolor{vermilion}{$\downarrow 13.00$}) & \textbf{45.00} (\textcolor{vermilion}{$\downarrow 23.00$}) \\
      & LLaVA-1.6 & 7B & 64.00 & 63.00 ($\downarrow 1.00$) & 64.00 ($\downarrow 0.00$) & \underline{43.00} ($\downarrow 21.00$) & 47.00 (\textcolor{vermilion}{$\downarrow 17.00$}) & \textbf{40.00} (\textcolor{vermilion}{$\downarrow 24.00$}) \\
      & OpenFlamingo & 3B & 45.00 & 39.00 ($\downarrow 6.00$) & 44.00 ($\downarrow 1.00$) & \underline{19.00} ($\downarrow 26.00$) & 20.00 (\textcolor{vermilion}{$\downarrow 25.00$}) & \textbf{14.00} (\textcolor{vermilion}{$\downarrow 31.00$}) \\
      & Blip-2 (FlanT5\textsubscript{XL}, ViT-L) & 3.4B & 63.00 & 58.00 ($\downarrow 5.00$) & 50.00 ($\downarrow 13.00$) & 38.00 ($\downarrow 25.00$) & \underline{36.00} (\textcolor{vermilion}{$\downarrow 27.00$}) & \textbf{32.00} (\textcolor{vermilion}{$\downarrow 31.00$}) \\
    \midrule
    \multirow{2}{*}{EVA-CLIP ViT-G/14}
      & Blip-2 (FlanT5\textsubscript{XL}) & 4.1B & 54.00 & 53.00 ($\downarrow 1.00$) & 54.00 ($\downarrow 0.00$) & 33.00 ($\downarrow 21.00$) & \underline{32.00} (\textcolor{vermilion}{$\downarrow 22.00$}) & \textbf{28.00} (\textcolor{vermilion}{$\downarrow 26.00$}) \\
      & InstructBLIP (FlanT5\textsubscript{XL}) & 4.1B & 68.00 & 64.00 ($\downarrow 4.00$) & 62.00 ($\downarrow 6.00$) & \underline{44.00} ($\downarrow 24.00$) & 45.00 (\textcolor{vermilion}{$\downarrow 23.00$}) & \textbf{41.00} (\textcolor{vermilion}{$\downarrow 27.00$}) \\
    \bottomrule
  \end{tabular}
\end{table*}

\begin{figure*}[t]
    \centering
    \includegraphics[width=\textwidth]{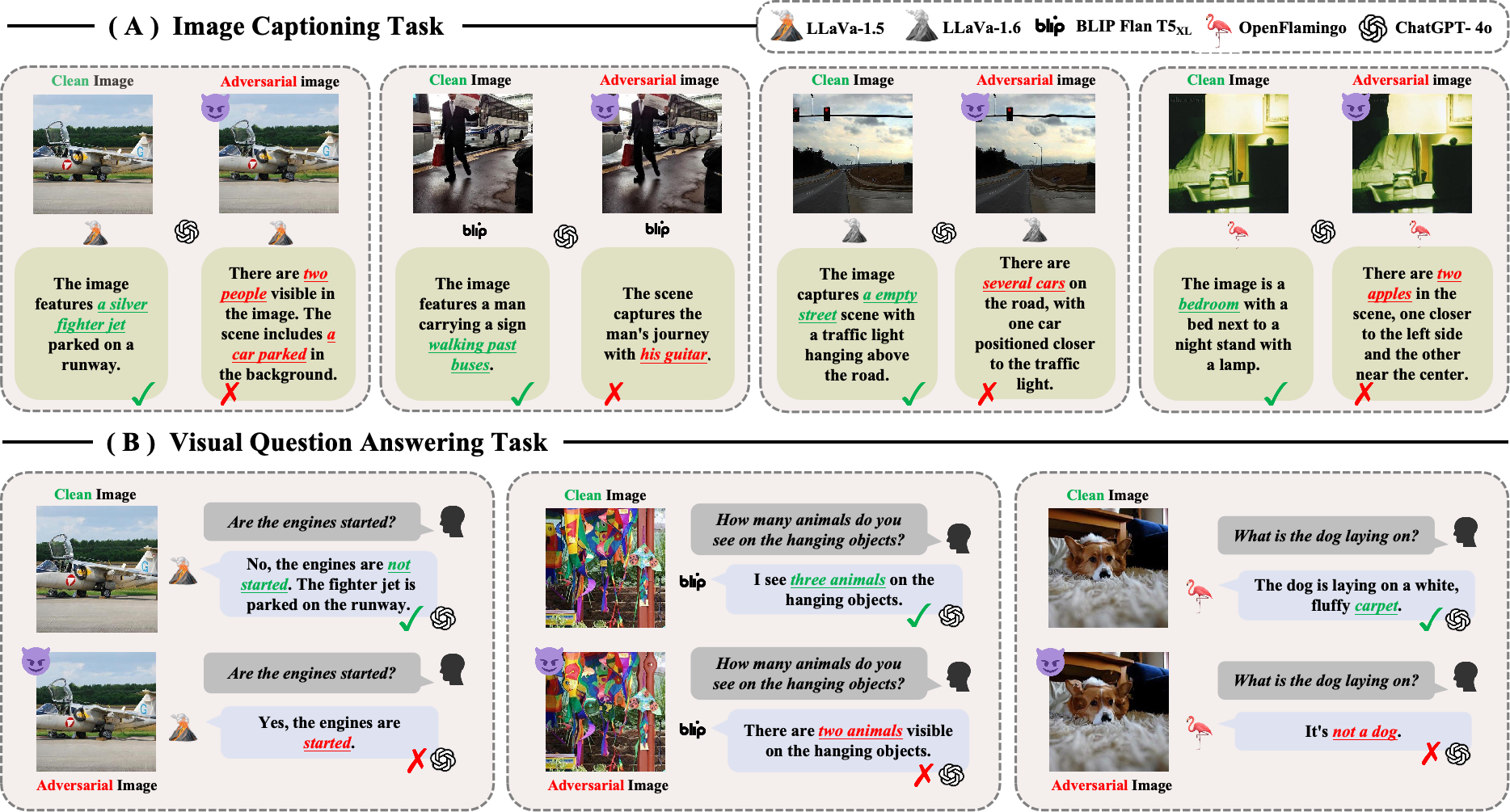}
    \caption{{\bfseries Qualitative Results on Image Captioning and Visual Question-Answering. }{\normalfont The figure shows representative outputs for clean and perturbed images, illustrating caption deviations and incorrect answers caused by the proposed perturbation.}}
    \Description{Qualitative results on image captioning and visual question-answering for clean and perturbed images.}
    \label{fig:VQA}
\end{figure*}

Figure~\ref{fig:figure2} (a) presents representative adversarial examples generated by the proposed method. Although the perturbations remain visually subtle and structurally coherent, they are sufficient to induce misclassification. Figure~\ref{fig:figure2} (b) further shows the corresponding attention visualizations under different CLIP models. Compared with clean images, the perturbed images shift model attention away from semantically relevant object regions toward less informative local textures or surrounding areas. This observation is consistent with the classification degradation, suggesting that the proposed perturbation influences both the final prediction and the intermediate attention distribution.

\textbf{Evaluation on the Image Captioning Task.} We transfer the adversarial examples generated in the zero-shot classification stage to the image captioning task to evaluate the cross-task transferability of the proposed wrinkle-like structural perturbation. As shown in Table~\ref{tab:transfer_zeroshot}, \emph{Ours (best)} achieves the lowest semantic consistency scores on all evaluated models, indicating the strongest transfer attack effect among the compared methods. The scores drop to 54.51, 53.15, 30.18, and 58.21 on LLaVA-1.5, LLaVA-1.6, OpenFlamingo, and Blip-2 (FlanT5\textsubscript{XL}, ViT-L), respectively, for perturbations generated from OpenAI CLIP ViT-L/14, and to 57.08 and 56.57 on Blip-2 (FlanT5\textsubscript{XL}) and InstructBLIP (FlanT5\textsubscript{XL}), respectively, for perturbations generated from EVA-CLIP ViT-G/14. Compared with Natural Light Attack~\cite{hsiao2024natural}, Shadow Attack~\cite{zhong2022shadows}, and ITA~\cite{liu2025lighting}, these results show that the proposed perturbation more effectively disrupts the alignment between visual content and generated descriptions. 

This tendency is also reflected in Figure~\ref{fig:VQA} (A). For clean images, the models generally describe the main scene correctly, such as a fighter jet on a runway, a man walking past buses, an empty street with a traffic light, or a bedroom with a bed and a lamp. After perturbation, the generated captions become misleading, often introducing non-existent objects or distorted scene semantics, such as a parked car in the fighter-jet image, a guitar in the bus scene, several cars on the empty road, or two apples in the bedroom scene.

\textbf{Evaluation on the VQA Task.} We further evaluate the generated adversarial examples on the visual question answering task. As reported in Table~\ref{tab:vqa}, \emph{Ours (best)} again obtains the lowest answer correctness scores on the evaluated models, indicating a consistently stronger transfer attack effect than Natural Light Attack~\cite{hsiao2024natural}, Shadow Attack~\cite{zhong2022shadows}, and ITA~\cite{liu2025lighting}. This further demonstrates effective transfer to downstream VQA reasoning. The correctness scores drop to 45.00, 40.00, 14.00, and 32.00 on LLaVA-1.5, LLaVA-1.6, OpenFlamingo, and Blip-2 (FlanT5\textsubscript{XL}, ViT-L), for perturbations generated from OpenAI CLIP ViT-L/14, and to 28.00 and 41.00 on Blip-2 (FlanT5\textsubscript{XL}) and InstructBLIP (FlanT5\textsubscript{XL}), respectively, for perturbations generated from EVA-CLIP ViT-G/14. Since VQA requires explicit reasoning, the substantial degradation suggests that the proposed perturbation interferes with both visual recognition and fine-grained cross-modal reasoning. 

This effect is also evident in Figure~\ref{fig:VQA} (B). For clean images, the models usually answer the questions correctly, such as recognizing that the fighter-jet engines are not started, counting three animals in the scene, or identifying that the dog is lying on a white carpet. After perturbation, the answers become incorrect or semantically inconsistent with the image, for example claiming that the engines are started, undercounting the animals as two, or responding that the object is not a dog. Together with the quantitative results, these examples show that the proposed perturbation effectively disrupts semantic reasoning in downstream VQA systems.

\section{Ablation Experiments}
To better understand the proposed method, we perform several ablation studies on its key design choices and search settings.

\begin{figure}[H]
    \centering
    \includegraphics[width=1\linewidth]{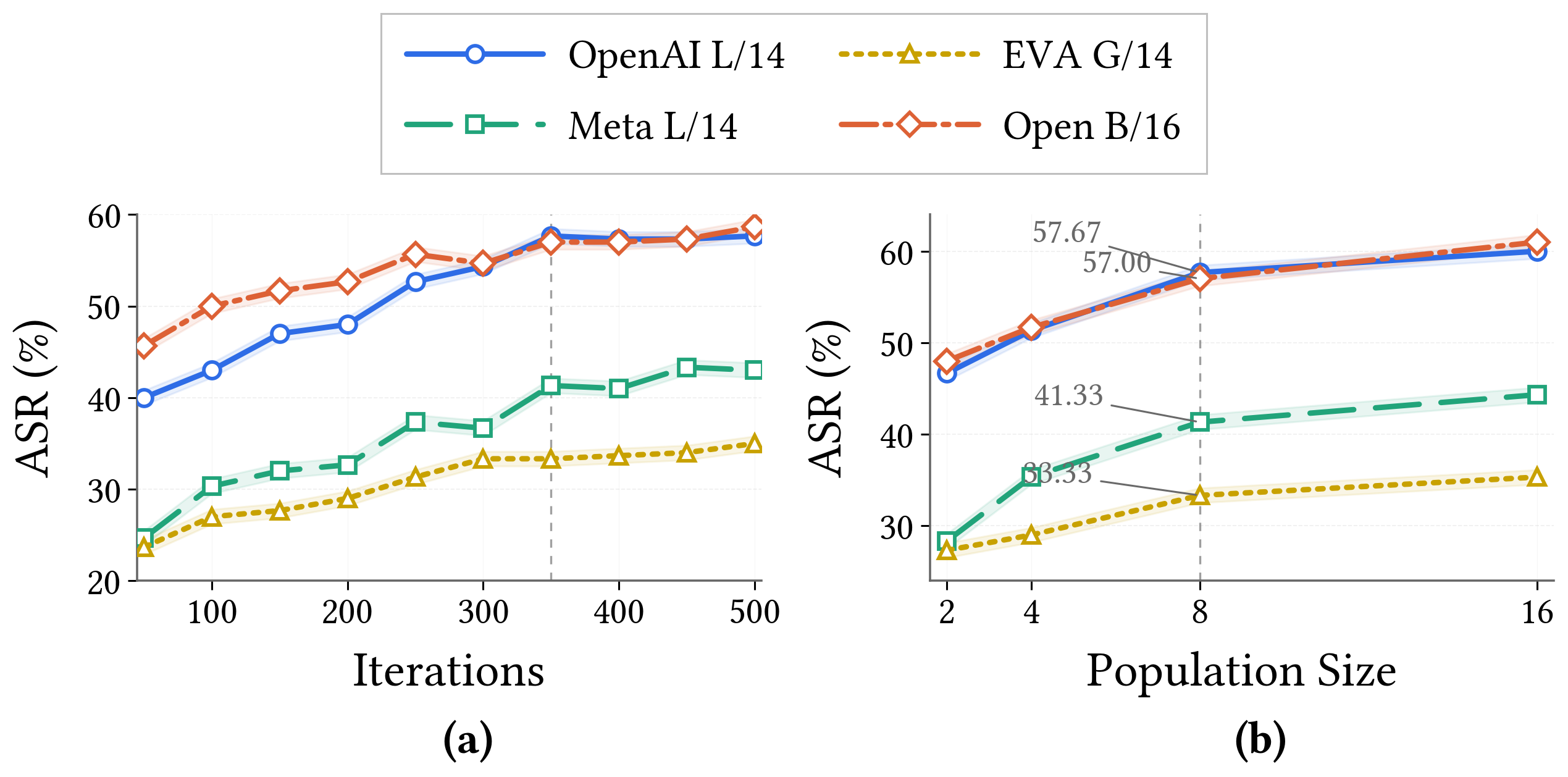}
    \caption{{\bfseries Ablation on Genetic Search Hyperparameters. }{\mdseries Attack performance of the proposed wrinkle-like structural perturbation under (a) different numbers of fitness evaluations and (b) different population sizes, showing the trade-off between attack effectiveness and search efficiency across different CLIP models.}}
    \label{fig:Ablation_1}
\end{figure}

\textbf{Effect of Fitness-Evaluation Budget.}
We first study the effect of the fitness-evaluation budget on attack performance. As shown in Figure~\ref{fig:Ablation_1} (a), increasing the number of fitness evaluations consistently improves ASR across different CLIP backbones, indicating more effective exploration of the parameter space. However, the gain gradually saturates near 350, suggesting limited additional benefit despite higher computational cost. This result supports the reasonableness of using 350 fitness evaluations in our setting.

\textbf{Effect of Population Size.}
We then study the effect of population size on attack effectiveness and search efficiency. As shown in Figure~\ref{fig:Ablation_1} (b), a small population size limits exploration, whereas an overly large population increases per-generation cost without proportional gains. Among the tested settings, a population size of 8 provides the best trade-off between attack effectiveness and efficiency, supporting the rationality of our parameter choice.

\begin{figure}[H]
    \centering
    \includegraphics[width=0.6\linewidth]{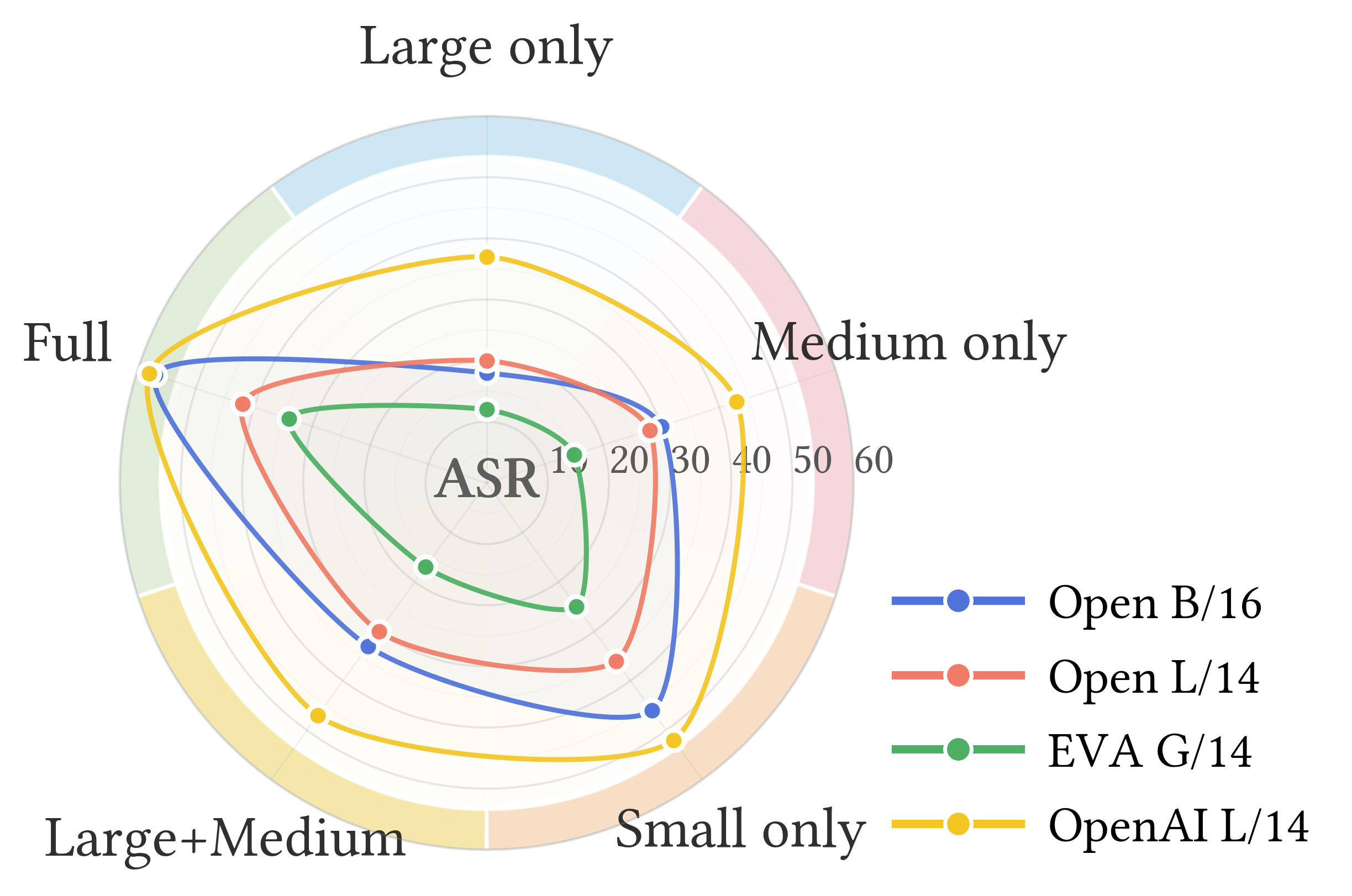}
    \caption{{\bfseries Ablation on multi-scale wrinkle components.} {\normalfont ASR (\%) under different component combinations is reported, including large only, medium only, small only, large+medium, and full.}}
    \label{fig:Ablation_2}
\end{figure}

\textbf{Effect of Multi-scale Wrinkle Components.}
We further analyze the effect of different wrinkle components on attack performance. Here, \textit{Full} combines large-, medium-, and small-scale perturbations; \textit{Large only}, \textit{Medium only}, and \textit{Small only} use only the corresponding single-scale component; and \textit{Large+Medium} uses the large- and medium-scale components without the small-scale one. As shown in Figure~\ref{fig:Ablation_2}, \textit{Full} achieves the best or near-best ASR across all models. Among the reduced settings, \textit{Small only} generally outperforms \textit{Large only} and \textit{Medium only}, while \textit{Large+Medium} still remains inferior to \textit{Full}. These results indicate that the proposed attack benefits from multi-scale complementarity, with the small-scale component playing an important role.

\begin{figure}[H]
    \centering
    \begin{minipage}[t]{0.49\linewidth}
        \centering
        \includegraphics[width=\linewidth]{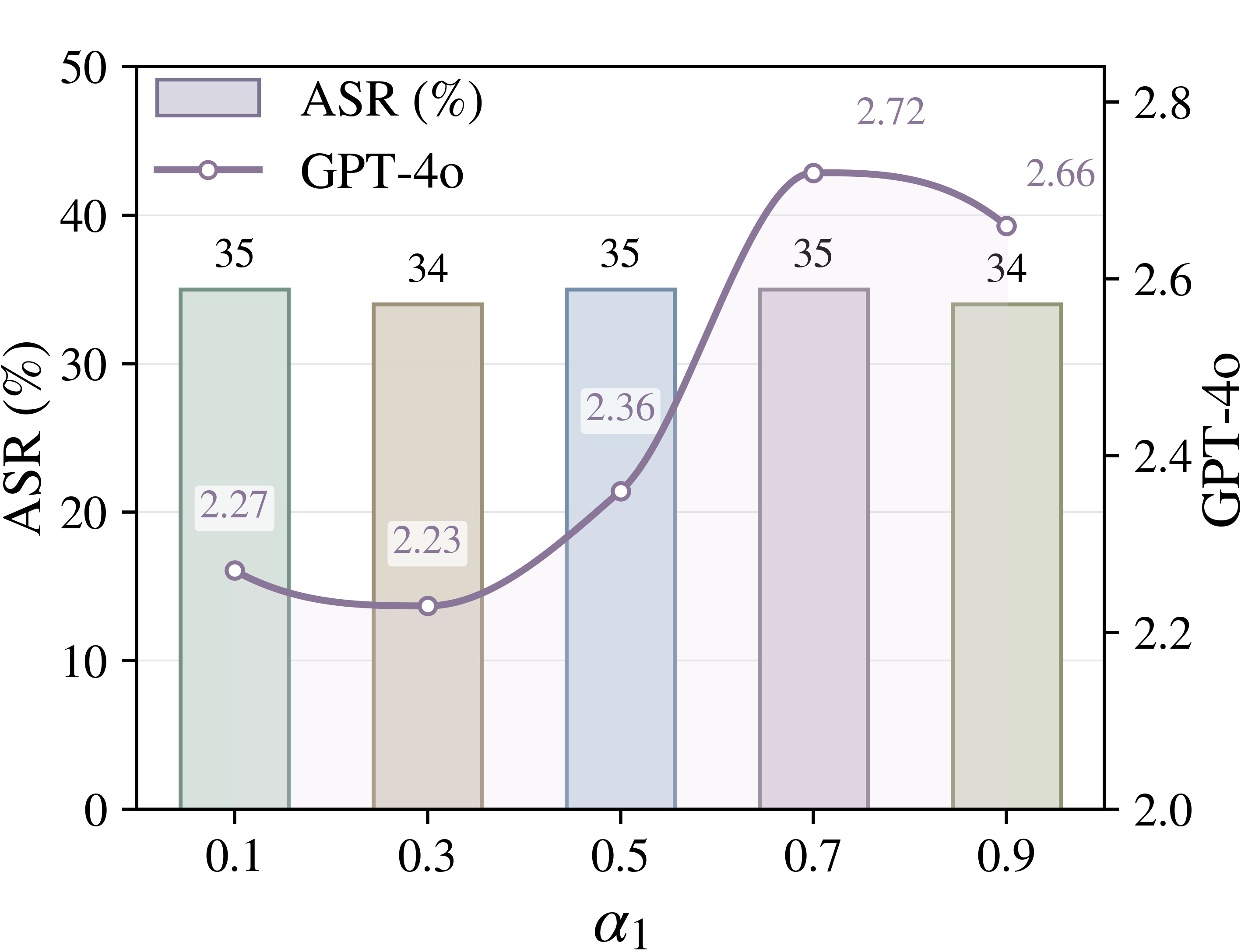}
        {\footnotesize (a)\par}
    \end{minipage}\hfill
    \begin{minipage}[t]{0.49\linewidth}
        \centering
        \includegraphics[width=\linewidth]{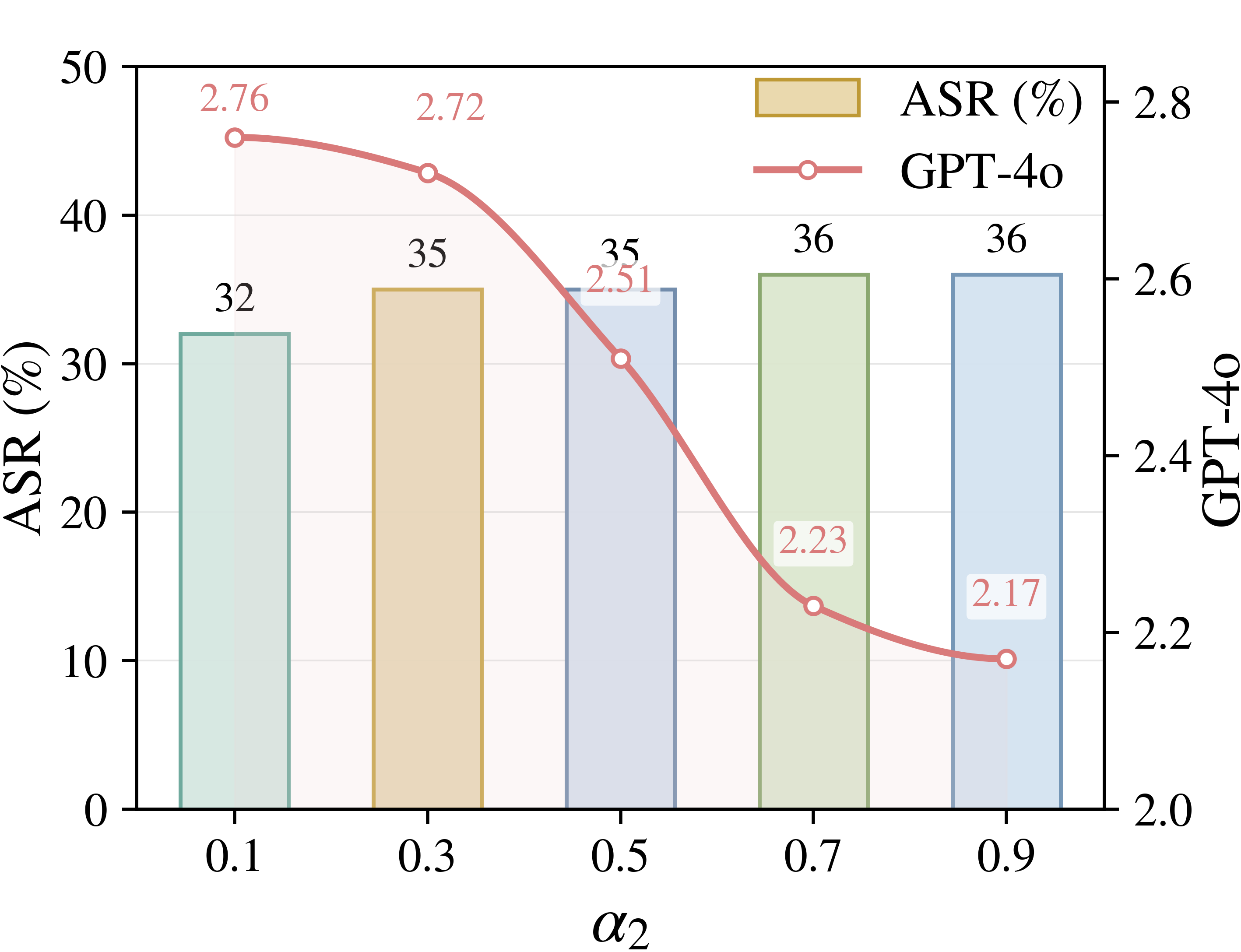}
        {\footnotesize (b)\par}
    \end{minipage}
    \caption{Ablation study of the weighting coefficients $\alpha_1$ and $\alpha_2$ on EVA-CLIP ViT-G/14.}
    \label{fig:ablation}
\end{figure}

\textbf{Effect of the Perceptual-Fusion Coefficient.}
Using EVA-CLIP ViT-G/14 as an example, we further study the effect of the perceptual-fusion coefficient $\alpha_1$. As shown in Figure~\ref{fig:ablation} (a), changing $\alpha_1$ causes limited variation in ASR, while the GPT-4o score changes more noticeably. In particular, $\alpha_1 = 0.7$ achieves the best visual quality while maintaining competitive attack performance. This indicates that assigning a higher weight to the LPIPS-based perceptual term yields a better balance between structural preservation and high-level perceptual consistency. Therefore, $\alpha_1 = 0.7$ is a reasonable choice.

\textbf{Effect of the Attack--Perceptual Weighting Coefficient.}
Using EVA-CLIP ViT-G/14 as an example, we study the effect of the attack--perceptual weighting coefficient $\alpha_2$ on attack performance and visual quality. As shown in Figure~\ref{fig:ablation} (b), a larger $\alpha_2$ generally improves ASR, while causing a clear drop in the GPT-4o score, indicating lower visual quality. This suggests that increasing the weight of the perceptual term improves attack effectiveness, but weakens the visual naturalness of the perturbations. Considering this trade-off, $\alpha_2 = 0.3$ is a reasonable setting, as it provides a balanced compromise between attack success and visual quality.

\section{Conclusion}
The robustness of vision-language models under wrinkle-like structural perturbations is systematically evaluated through a parametric attack framework based on non-rigid geometric deformation. Experimental results show consistent performance degradation in zero-shot classification, together with effective transfer to image captioning and visual question answering. Compared with the baseline methods, the proposed method demonstrates stronger attack effectiveness across multiple models, while attention visualizations indicate that the perturbation shifts model focus away from semantically relevant regions. These findings suggest that vision-language models remain vulnerable to structured non-rigid perturbations beyond additive noise.

\bibliographystyle{ACM-Reference-Format}
\bibliography{refs_acmmm-4_clean}

\end{document}